\documentclass[10pt,twocolumn,letterpaper]{article}

\usepackage{cvpr}
\usepackage{times}
\usepackage{epsfig}
\usepackage{graphicx}
\usepackage{amsmath}
\usepackage{amssymb}
\usepackage{colortbl}
\usepackage[ruled,vlined]{algorithm2e}
\usepackage{algorithmic}
\usepackage{subfigure}
\usepackage{breqn}
\definecolor{mygray}{gray}{.7}
\usepackage{footnote}
\makeatletter
\def\hlinew#1{%
  \noalign{\ifnum0=`}\fi\hrule \@height #1 \futurelet
   \reserved@a\@xhline}
\makeatother

\def\figvspace{{\vspace{-2mm}}}

\usepackage[pagebackref=true,breaklinks=true,letterpaper=true,colorlinks,bookmarks=false]{hyperref}

\cvprfinalcopy 

\def\cvprPaperID{5450} 

\ifcvprfinal\fi
\begin{document}

\title{CurricularFace: Adaptive Curriculum Learning Loss for Deep Face Recognition}

\author{Yuge Huang$^{\dag}$ ~ ~ Yuhan Wang$^{\S}$ ~ ~ Ying Tai$^{\dag}$\thanks{~denotes Ying Tai and Shaoxin Li are corresponding authors.} ~ ~ Xiaoming Liu$^{\ddag}$\\
 ~ ~ Pengcheng Shen$^{\dag}$~ ~ Shaoxin Li$^{\dag \ast}$~ ~ Jilin Li$^{\dag}$  ~ ~ Feiyue Huang$^{\dag}$\\
$^\dag$Youtu Lab, Tencent  ~ ~ ~  $^\S$Zhejiang University ~ ~ ~ $^\ddag$Michigan State University \\
{\tt\small $^\dag$\{yugehuang, yingtai, quantshen, darwinli, jerolinli, garyhuang\}@tencent.com} \\
{\tt\small $^\S$wang_yuhan@zju.edu.cn, $^\ddag$liuxm@cse.msu.edu}\\
{\small \url{https://github.com/HuangYG123/CurricularFace}}
}

\maketitle

\begin{abstract}
As an emerging topic in face recognition, designing margin-based loss functions can increase the feature margin between different classes for enhanced discriminability.
More recently, the idea of mining-based strategies is adopted to emphasize the misclassified samples, achieving promising results.
However, during the entire training process, the prior methods either do not explicitly emphasize the sample based on its importance that renders the hard samples not fully exploited; or explicitly emphasize the effects of semi-hard/hard samples even at the early training stage that may lead to convergence issue.
In this work, we propose a novel Adaptive Curriculum Learning loss (CurricularFace) that embeds the idea of curriculum learning into the loss function to achieve a novel training strategy for deep face recognition, which mainly addresses easy samples in the early training stage and hard ones in the later stage.
Specifically, our CurricularFace adaptively adjusts the relative importance of easy and hard samples during different training stages.
In each stage, different samples are assigned with different importance according to their corresponding difficultness.
Extensive experimental results on popular benchmarks demonstrate the superiority of our CurricularFace over the state-of-the-art competitors.
\end{abstract}

\begin{figure}[t]
  \centering
  \includegraphics[trim={0 0 0 0mm},clip,width=1\linewidth]{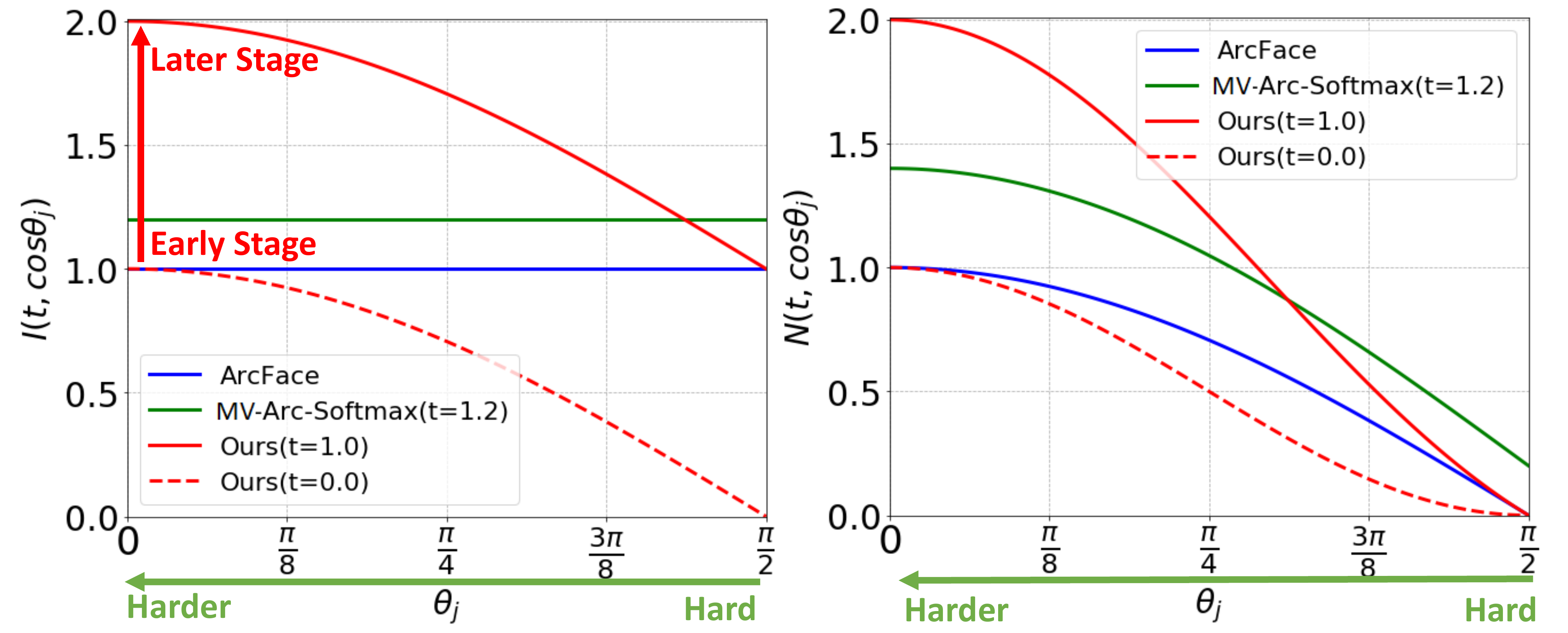}
  \caption{\small
  \textbf{Different training strategies} for modulating negative cosine similarities of hard samples (\textit{i.e.}, the mis-classified samples) in ArcFace~\cite{deng2018arcface}, MV-Arc-Softmax~\cite{wang2018support} and our CurricularFace.
  \textbf{Left}: The modulation coefficients $I(t,\cos\theta_j)$ for negative cosine similarities of hard samples in different methods, where $t$ is an adaptively estimated parameter and $\theta_j$ denotes the angle between the hard sample and the non-ground truth $j$-class center.
  \textbf{Right}: The corresponding hard samples' negative cosine similarities $N(t,\cos\theta_j) = I(t,\cos\theta_j)\cos\theta_j + c$ after modulation, where $c$ indicates a constant.
  On one hand, during early training stage (\textit{e.g.}, $t$ is close to $0$), hard sample's negative cosine similarities are usually reduced, and thus leads to smaller hard sample loss than the original one. Therefore, easier samples are relatively emphasized; during later training stage (\textit{e.g.}, $t$ is close to $1$), the hard sample's negative cosine similarities are enhanced, and thus leads to larger hard sample loss.
  On the other hand, in the same training stage, we modulate the hard samples' negative cosine similarities with $\cos\theta_j$.
  Specifically, \textbf{the smaller the angle $\theta_j$ is, the larger the modulation coefficient should be}.
  }
  \label{fig:hard_sample_weight}
  \figvspace
\end{figure}

\section{Introduction}
The success of Convolutional Neural Networks (CNNs) on face recognition can be mainly credited to: enormous training data, network architectures, and loss functions.
Recently, designing effective loss functions that enhance discriminative power is pivotal for training deep face CNNs.

Current state-of-the-art (SOTA) face recognition methods mainly adopt softmax-based classification loss.
Since the learned features with the original softmax is not sufficiently discriminative for the practical face recognition problem~\cite{liu2017sphereface}, which means that the testing identities are usually disjoint from the training set, several margin-based variants have been proposed to enhance features' discriminative power.
For example, explicit margin, \textit{i.e.},  CosFace~\cite{wang2018cosface}, Sphereface~\cite{liu2017sphereface}, ArcFace~\cite{deng2018arcface}, and implicit margin, \textit{i.e.}, Adacos~\cite{zhang2019adacos}, supplement the original softmax function to enforce greater intra-class compactness and inter-class discrepancy, which result in more discriminate features.
However, these margin-based loss functions \textit{do not explicitly emphasize each sample according to its importance}.

As demonstrated in~\cite{chen2019angular,huang2020distribution}, hard sample mining is also a critical step to further improve the final accuracy.
As a commonly-used hard sample mining method, OHEM~\cite{shrivastava2016training} focuses on the large-loss samples in one mini-batch, in which the percentage of hard samples is empirically decided and easy samples are completely discarded.
Focal loss~\cite{lin2017focal} is a soft mining variant that rectifies the loss function to a elaborately designed form, in which two hyper-parameters should be tuned with a lot of efforts to decide the weights of each sample and hard samples are emphasized by reducing the weights of easy samples.
Recently, Triplet loss~\cite{schroff2015facenet} and MV-Arc-Softmax~\cite{wang2018support} are motivated by integrating both margin and mining into one framework.
Triplet loss adopts a semi-hard mining strategy to obtain semi-hard triplets and enlarges the margin between triplet samples.
MV-Arc-Softmax~\cite{wang2018support} clearly defines hard samples as misclassified samples and emphasizes them by increasing the weights of their negative cosine similarities with a preset constant.
In a nutshell, mining-based loss functions \textit{explicitly emphasize the effects of semi-hard or hard samples~\cite{schroff2015facenet}}.

However, there are drawbacks in training strategies of both margin- and mining-based loss functions. The general softmax-based loss function can be formulated as follows:
\begin{equation}
\small{}
\label{eq:general_softmax_first}
{
\mathcal{L} = -\log\frac{e^{sT(\cos\theta_{y_i})}}
{e^{s{T(\cos\theta_{y_i})}+\sum^{n}_{j=1,j\neq y_{i}}
e^{sN(t, \cos\theta_{j})}},
}
}
\end{equation}
where $T(\cos\theta_{y_i})$ and $N(t,\cos\theta_{j}) = I(t, \cos\theta_j)\cos\theta_j+c$ are the functions to define the positive and negative cosine similarities, respectively. $I(t, \cos\theta_j)$ denotes the modulation coefficients of negative cosine similarities and  $c$ is a constant.
For margin-based methods, mining strategy is ignored and thus the difficultness of each sample is not exploited, which may lead to convergence issues when using a large margin on small backbones, \textit{e.g.}, MobileFaceNet~\cite{chen2018mobilefacenets}.
As shown in Fig.~\ref{fig:hard_sample_weight}, the modulation coefficients $I(\cdot)$ for the negative cosine similarities are fixed as a constant of $1$ in ArcFace for all samples during the entire training process.
For mining-based methods, over-emphasizing hard samples in early training stage may hinder the model to converge. MV-Arc-Softmax emphasizes hard samples by modulating the negative cosine similarity as $N(t,\cos\theta_{j})=t\cos\theta_{j}+t-1$, \textit{i.e.}, $I(t,\cos\theta_j) = t$, where $t$ is a manually defined constant.
As MV-Arc-Softmax claimed, $t$ plays a key role in the model convergence property and a slight larger value (\textit{e.g.}, $>$$1.4$) may cause the model difficult to converge.
Thus $t$ needs to be carefully tuned.

In this work, we propose a novel adaptive curriculum learning loss, termed CurricularFace, to achieve a novel training strategy for deep face recognition.
Motivated by the nature of human learning that easy cases are learned first and then come the hard ones~\cite{bengio2009curriculum}, our {CurricularFace} incorporates the idea of Curriculum Learning (CL) into face recognition in an \textit{adaptive} manner, which differs from the traditional CL in two aspects.
First, \textbf{the curriculum construction is adaptive}.
In traditional CL, the samples are \textit{ordered} by the corresponding difficultness, which are often defined by a prior and then fixed to establish the curriculum.
In CurricularFace, the samples are \textit{randomly} selected in each mini-batch, while the curriculum is established adaptively via mining the hard samples online, which shows the \textit{diversity} in samples with different importance.
Second, \textbf{the importance of hard samples are adaptive}.
On one hand, the relative importance between easy and hard samples is dynamic and could be adjusted in different training stages.
On the other hand, the importance of each hard sample in current mini-batch depends on its own difficultness.

Specifically, the mis-classified samples in mini-batch are chosen as hard samples and weighted by adjusting the modulation coefficients $I(t,cos\theta_j)$ of cosine similarities between the sample and the non-ground truth class center vectors, \textit{i.e.}, negative cosine similarity $cos\theta_j$.
To achieve the goal of adaptive curricular learning in the entire training, we design a novel coefficient function $I(\cdot)$ that is determined by two factors:
$1$) the adaptively estimated parameter $t$ that utilizes moving average of positive cosine similarities between samples and the corresponding ground-truth class center
to unleash the burden of manually tuning;
and $2$) the angle $\theta_j$ that defines the difficultness of hard samples to achieve adaptive assignment.
To sum up, the contributions of this work are: 
\begin{itemize}
\item We propose an adaptive curriculum learning loss for face recognition, which automatically emphasizes easy samples first and hard samples later.
To the best of our knowledge, it is the first work to introduce the idea of adaptive curriculum learning for face recognition.
\item We design a novel modulation coefficient function $I(\cdot)$ to achieve adaptive curriculum learning during training, which connects positive and negative cosine similarity simultaneously without the need of manually tuning any additional hyper-parameter.
\item We conduct extensive experiments on popular facial benchmarks, 
which demonstrate the superiority of our CurricularFace over the SOTA competitors.
\end{itemize}

\section{Related Work}
\label{gen_inst}
\paragraph{Margin-based loss function.}
Loss design is pivotal for large-scale face recognition.
Current SOTA deep face recognition methods mostly adopt softmax-based classification loss~\cite{taigman2014deepface}.
Since the learned features with the original softmax loss are not guaranteed to be discriminative enough for practical face recognition problem~\cite{liu2017sphereface}, margin-based losses~\cite{liu2016large,liu2017sphereface, deng2018arcface} are proposed.
Though the margin-based loss functions are verified to obtain good performance, they do not take the difficultness of each sample into consideration, while our CurricularFace emphasizes easy samples first and hard samples later, which is more reasonable and effective.

\paragraph{Mining-based loss function.}
Though some mining-based loss function such as Focal loss~\cite{lin2017focal}, Online Hard Sample Mining (OHEM)~\cite{shrivastava2016training} are prevalent in the field of object detection, they are rarely used in face recognition.
OHEM focuses on the large-loss samples in one mini-batch, in which the percentage of the hard samples is empirically determined and easy samples are completely discarded. 
Focal loss emphasizes hard samples by reducing the weights of easy samples, in which two hyper-parameters should be manually tuned.
The recent work, MV-Arc-Softmax~\cite{wang2018support} fuses the motivations of both margin and mining into one framework for deep face recognition. They define hard samples as misclassified samples and enlarge the weights of hard samples with a preset constant.
Our method differs from MV-Arc-Softmax in three aspects:
$1$) We do not always emphasize hard samples, especially in the early training stages.
$2$) We assign different weights for hard samples according to their corresponding difficultness.
$3$) We adaptively estimate the additional hyper-parameter t without manual tuning.

\paragraph{Curriculum Learning.}
Learning from easier samples first and harder samples later is a common strategy in Curriculum Learning (CL)~\cite{bengio2009curriculum,zhou2018scheduled}.
The key problem in CL is to define the difficultness of each sample.
For example,~\cite{basu2013teaching} takes the negative distance to the boundary as the indicator for easiness in classification.
However, the ad-hoc curriculum design in CL turns out to be difficult to implement in different problems.
To alleviate this issue,~\cite{kumar2010selfpace} designs a new formulation, called Self-Paced Learning (SPL), where examples with lower losses are considered to be easier and emphasized during training.
The key differences between our CurricularFace with SPL are:
$1$) Our method focuses on easier samples in the early training stage and emphasizes hard samples in the later stage.
$2$) Our method proposes a novel function $N(\cdot)$ for negative cosine similarities, which achieves not only adaptive assignment on modulation coefficients $I(\cdot)$ for different samples in the same training stage,
but also adaptive curriculum learning strategy in different stages.

\section{The Proposed CurricularFace}
\subsection{Preliminary Knowledge on Loss Function}
The original softmax loss is formulated as follows:
\begin{equation}
\small{}
\label{eq:softmax}
{
\mathcal{L} = -\log\frac{e^{W_{y_{i}}x_{i}+b_{y_{i}}}}{\sum^{n}_{j=1}e^{W_{j}x_{i}+b_{j}}},
}
\end{equation}
where $x_{i}\in R^d$ denotes the deep feature of $i$-th sample which belongs to the $y_{i}$ class, $W_{j} \in R^d$ denotes the $j$-th column of the weight $W\in R^{d\times n}$ and $b_j$ is the bias term.
The class number and the embedding feature size are $n$ and $d$, respectively.
In practice, the bias is usually set to $b_{j} = 0$ and the individual weight is set to $||W_{j}||=1$ by $l_2$ normalization. The deep feature is also normalized and re-scaled to $s$. Thus, the original softmax can be modified as follows:
\begin{equation}
\small{}
\label{eq:softmax_modify}
{
\mathcal{L} = -\log\frac{e^{s(\cos\theta_{y_{i}})}}{e^{s(\cos\theta_{y_{i}})}+\sum^{n}_{j=1,j\neq y_{i}}e^{s(\cos\theta_{j})}}.
}
\end{equation}
Since the learned features with original softmax loss may not be discriminative enough for practical face recognition problem, several variants are proposed and can be formulated in a general form:
\begin{equation}
\small{}
\label{eq:general_softmax}
{
\mathcal{L} = -G(p(x_i))\log\frac{e^{sT(\cos\theta_{y_i})}}
{e^{s{T(\cos\theta_{y_i})}}+\sum^{n}_{j=1,j\neq y_{i}}
{e^{sN(t,\cos\theta_{j})}},
}
}
\end{equation}
where $p(x_i)=\frac{e^{sT(\cos\theta_{y_i})}}{e^{s{T(\cos\theta_{y_i})}}+{\sum^{n}_{j=1,j\neq y_{i}}
e^{sN(t, \cos\theta_{j})}}}$ is the predicted ground truth probability and $G(p(x_i))$ is an indicator function.
$T(\cos\theta_{y_i})$ and $N(t,\cos\theta_{j}) = I(t, \cos\theta_j)\cos\theta_j+c$ are the functions to modulate the positive and negative cosine similarities, respectively, where $c$ is a constant, and $I(t, \cos\theta_j)$ denotes the modulation coefficients of negative cosine similarities.
In margin-based loss function, \textit{e.g.}, ArcFace, $G(p(x_i))=1$, $T(\cos\theta_{y_i})=\cos(\theta_{y_i}+m)$, and $N(t, \cos\theta_{j})=\cos\theta_{j}$.
It only modifies the positive cosine similarity of each sample to enhance the feature discrimination. As shown in Fig.~\ref{fig:hard_sample_weight}, the modulation coefficients $I(\cdot)$ of each sample's negative cosine similarities are fixed as $1$.
The recent work, MV-Arc-Softmax emphasizes hard samples by increasing $I(t,\cos\theta_{j})$ for hard samples.
That is, $G(p(x_i))=1$ and $N(t, \cos{\theta_{j}})$ is formulated as follows:
\begin{equation}
\small{}
\label{eq:sv_softmax}
N(t, cos_{\theta_{j}}) =
\begin{cases}
\cos\theta_{j},&\mbox{$T(\cos\theta_{y_{i}})-\cos\theta_j \geq 0$} \\
t\cos\theta_{j}+t-1,&\mbox{$T(\cos\theta_{y_{i}})-\cos\theta_j < 0$}.
\end{cases}
\end{equation}
If a sample is defined to be easy, its negative cosine similarity is kept the same as the original one, $\cos\theta_j$; if as a hard sample, its negative cosine similarity becomes $t\cos\theta_j+t-1$. That is, as shown in Fig.~\ref{fig:hard_sample_weight}, $I(\cdot)$ is a constant and determined by a preset hyper-parameter $t$. Meanwhile, since $t$ is always larger than $1$, $t\cos\theta_j+t-1 > \cos\theta_j$ always holds true, which means the model always focuses on hard samples, even in the early training stage.
However, the parameter $t$ is sensitive that a large pre-defined value (\textit{e.g.}, $>1.4$) may lead to convergence issue.

\subsection{Adaptive Curricular Learning Loss}
Next, we present the details of our proposed adaptive curriculum learning loss, which is the first attempt to introduce adaptive curriculum learning into deep face recognition.
The formulation of our loss function is also contained in the general form, where $G(p(x_i))=1$, positive and negative cosine similarity functions are defined as follows:
\begin{equation}
\small{}
\label{eq:ours_t}
{
T(\cos\theta_{y_i})=\cos(\theta_{y_i}+m),
}
\end{equation}
\begin{equation}
\small{}
\label{eq:ours_n}
N(t, \cos_{\theta_{j}}) =
\begin{cases}
\cos\theta_{j},&\mbox{$T(\cos\theta_{y_{i}})-\cos\theta_j \geq 0$} \\
\cos\theta_{j}(t+\cos\theta_{j}), &\mbox{$T(\cos\theta_{y_{i}})-\cos\theta_j<0$}.
\end{cases}
\end{equation}
It should be noted that the positive cosine similarity can adopt any margin-based loss functions and here we adopt ArcFace as an example.
As shown in Fig.~\ref{fig:hard_sample_weight}, the modulation coefficient $I(t,\theta_j)$ of hard sample negative cosine similarity depends on both the value of $t$ and $\theta_j$.
In the early training stage, learning from easy samples is beneficial to model convergence.
Thus, $t$ should be close to zero and $I(\cdot) = t+\cos\theta_{j}$ is smaller than $1$. 
Therefore, the weights of hard samples are reduced and easy samples are emphasized relatively.
As training goes on, the model gradually focuses on the hard samples, \textit{i.e.}, the value of $t$ shall increase and $I(\cdot)$ is larger than $1$.
Thus, the hard samples are emphasized with larger weights.
Moreover, within the same training stage, $I(\cdot)$ is monotonically decreasing with $\theta_{j}$ so that harder sample can be assigned with larger coefficient according to its difficultness.
The value of the parameter $t$ is \textit{automatically} estimated in our CurricularFace, 
otherwise it may require lots of efforts for manual tuning.

\begin{algorithm}[t]
\small
\SetAlgoLined
\KwIn{The deep feature of $i$-th sample $x_i$ with its label $y_i$, last fully-connected layer parameters $W$, cosine similarity $\cos\theta_j$ of two vectors, embedding network parameters $\Theta$, learning rate $\lambda$, and margin $m$}
iteration  number $k\leftarrow 0$, parameter $t\leftarrow 0$, $m\leftarrow 0.5$\;

 \While{not converged}{
  \eIf{$\cos(\theta_{y_i}+m) \geq \cos\theta_j$}{
   $N(t, \cos\theta_j)=\cos\theta_j$\;
   }{
   $N(t, \cos\theta_j)=(t^{(k)}+\cos\theta_j)\cos\theta_j$\ ;
  }
  $T(\cos\theta_{y_i}) = \cos(\theta_{y_i}+m)$\;
  Compute the loss $\mathcal{L}$ by Eq.~$\ref{eq:our_loss}$\;
  Compute the gradients of $x_i$ and $W_j$ by Eq.~$\ref{eq:ours_gradient_w_x}$\;
  Update the parameters $W$ and $\Theta$ by:
  $W^{(k+1)} = W^{(k)} - \lambda^{(k)}\frac{\partial L}{\partial W}$,
  $\Theta^{(k+1)} = \Theta^{(k)} - \lambda^{(k)}\frac{\partial L}{\partial x_i}\frac{\partial x_i}{\partial \Theta^{(k)}}$\;
  $k \leftarrow k+1$\;
  Update the parameter $t$ by Eq.~$\ref{eq:t_estimation}$\;
 }
\KwOut{$W$, $\Theta$.}
 \caption{CurricularFace}
 \label{alg:training}
\end{algorithm}

\begin{figure}[t!]
  \centering
  \includegraphics[trim={0 0 0 0mm},clip,width=\linewidth]{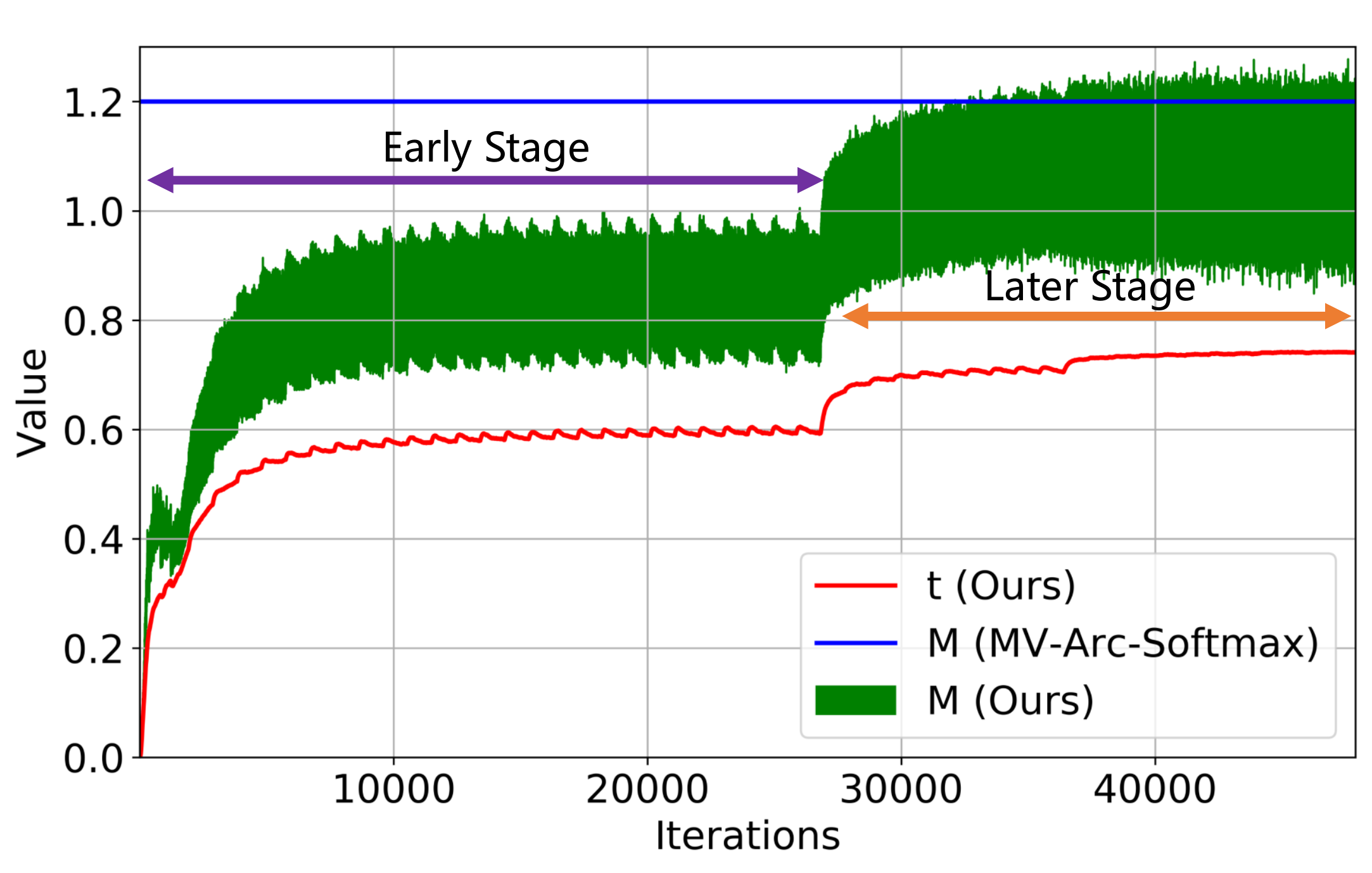}
   \caption{\small \textbf{The adaptive parameter $t$ (\textcolor{red}{red line}) and gradient modulation coefficients $M$ of ours (\textcolor{green}{green area}) and MV-Arc-Softmax (\textcolor{blue}{blue line}) in training}. Since the number of mined hard samples reduces as training progresses, the green area, {\it i.e.}, the range of $M$ values, is relatively smooth in early stage and exhibits burrs in later stage.
   }
    \label{fig:estimation_of_t}
    \figvspace
\end{figure}


\paragraph{Optimization.}
Next, we show our CurricularFace can be easily optimized by the conventional stochastic gradient descent.
Assuming $x_i$ denotes the deep feature of $i$-th sample which belongs to the $y_i$ class, the input of the proposed function is the logit $f_j$, where $j$ denotes the $j$-th class.

In the forwarding process, when $j=y_{i}$, it is the same as the ArcFace, \textit{i.e.}, $f_j=sT(\cos\theta_{y_{i}})$, $T(\cos\theta_{y_{i}})=\cos(\theta_{y_{i}}+m)$. When $j\neq y_i$, it has two cases, if $x_i$ is an easy sample, it is the the same as the original softmax, \textit{i.e.}, $f_j=s\cos\theta_{j}$. Otherwise, it will be modulated as $f_j=sN(t,\cos\theta_j)$, where $N(t,\cos\theta_j)=(t+\cos\theta_j)\cos\theta_j$.
In the backward propagation process, the gradients w.r.t. $x_i$ and $W_j$ can also be divided into three cases and computed as follows:
\begin{equation}
\small{}
\label{eq:ours_gradient_w_x}
\begin{aligned}
&\frac{\partial L}{\partial x_i} =
\begin{cases}
\frac{\partial L}{\partial f_{y_i}}(s\frac{\sin(\theta_{y_i}+m)} {\sin\theta_{y_i}})W_{y_{i}}, &\mbox{$j=y_{i}$}\\
\frac{\partial L}{\partial f_j}s W_j, &\mbox{$j \neq y_i$, easy} \\
\frac{\partial L}{\partial f_j}s(2\cos\theta_j+t)W_j &\mbox{$j \neq y_i$, hard}
\end{cases}\\
&\frac{\partial L}{\partial W_j} =
\begin{cases}
\frac{\partial L}{\partial f_{y_i}}(s\frac{\sin(\theta_{y_i}+m)} {\sin\theta_{y_i}})x_i, &\mbox{$j=y_{i}$}\\
\frac{\partial L}{\partial f_j}s x_i, &\mbox{$j \neq y_i$, easy} \\
\frac{\partial L}{\partial f_j}s(2\cos\theta_j+t)x_i &\mbox{$j \neq y_i$, hard}
\end{cases}
\end{aligned}
\end{equation}
Based on the above formulations, we can find the gradient modulation coefficients of hard samples are determined by $M(\cdot)=2\cos\theta_j+t$, which consists of two parts, the negative cosine similarity $\cos\theta_j$ and the value of $t$. As shown in Fig.~\ref{fig:estimation_of_t}, on the one hand, the coefficients increase with the adaptive estimation of $t$ (described in the next subsection) to emphasize hard samples. On the other hand, these coefficients are assigned with different importance according to their corresponding difficultness ($\cos \theta_j)$. Therefore, the values of $M$ in Fig.~\ref{fig:estimation_of_t} are plotted as a range at each training iteration. However, the coefficients are fixed to be $1$ and a constant $t$ in ArcFace and MV-Arc-Softmax, respectively.

\paragraph{Adaptive Estimation of $t$.}
It is critical to determine appropriate values of $t$ in different training stages.
Ideally the value of $t$ can indicate the model training stages.
We empirically find the \textit{average} of positive cosine similarities is a good indicator.
However, mini-batch statistic-based methods usually face an issue: when many extreme data are sampled in one mini-batch, the statistics can be vastly noisy and the estimation will be unstable.
Exponential Moving Average (EMA) is a common solution to address this issue~\cite{li2019gradient}.
Specifically, let $r^{(k)}$ be the \textit{average of the positive cosine similarities} of the $k$-th batch and be formulated as $r^{(k)}=\sum_{i}{\cos\theta_{y_{i}}}$, we have:
\begin{equation}
\small{}
\label{eq:t_estimation}
t^{(k)} = \alpha r^{(k)} + (1 - \alpha)t^{(k-1)},
\end{equation}
where $t^{0}=0$, $\alpha$ is the momentum parameter and set to $0.99$.
With the EMA, we avoid the hyper-parameter tuning and make the modulation coefficients of hard sample negative cosine similarities $I(\cdot)$ adaptive to the current training stage.
To sum up, the loss function of our CurricularFace is formulated as follows:
\begin{equation}
\small{}
\label{eq:our_loss}
{
\mathcal{L} = -\log\frac{e^{s\cos(\theta_{y_i}+m)}}
{e^{s\cos(\theta_{y_i}+m)}+\sum^{n}_{j=1,j\neq y_{i}}
e^{sN(t^{(k)}, \cos\theta_{j})}},
}
\end{equation}
where $N(t^{(k)}, \cos\theta_{j})$ is defined in Eq.~\ref{eq:ours_n}.
The entire training process is summarized in Algorithm~\ref{alg:training}.

\begin{figure}[t]
 \centering
 \includegraphics[trim={0 0 0 0mm},clip,width=0.92\linewidth]{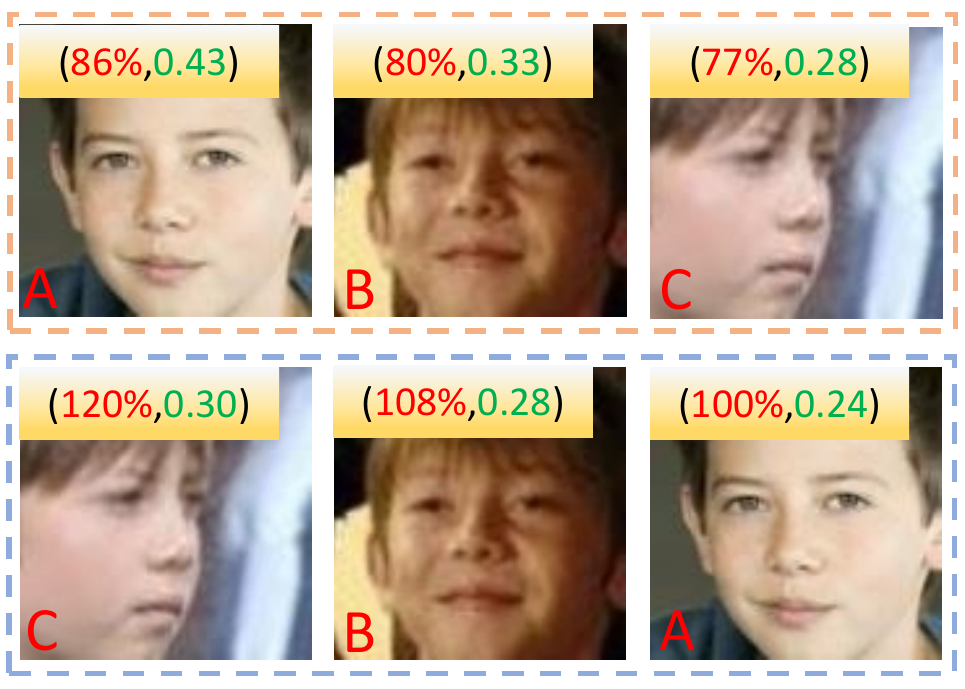}
 \caption{\small
 Illustrations on \textbf{(\textcolor{red}{ratio between our loss and ArcFace}, \textcolor{green}{maximum $cos\theta_j$})} in different training stages.
 \textbf{Top}: Early training stage.
 \textbf{Bottom}: Later training stage.
 }
 \label{fig:visualization}
 \figvspace
\end{figure}

Fig.~\ref{fig:visualization} illustrates how the loss changes from ArcFace to our CurricularFace during training.
Here are some observations:
$1$) As we excepted, hard samples (B and C) are suppressed in early training stage but emphasized later.
$2$) The ratio is monotonically increasing with $cos\theta_j$, since the larger $cos\theta_j$ is, the harder the sample is.
$3$) The positive cosine similarity of a perceptual-well image is often large.
However, during the early training stage, the negative cosine similarities of the perceptual-well image (A) may  also be large so that it could be classified as the hard one.


\begin{table}[t!]
\begin{center}
\small
\caption{\small \textbf{The decision boundaries of popular loss functions}.}
\label{tab:decision_boundary}
\resizebox{1\columnwidth}{!}{
\begin{tabular}{l|c}
\hlinew{1.2pt}
Loss           & Decision Boundary \\ \hline\hline
Softmax        &  $\cos\theta_{y_i} = \cos\theta_j$  \\ \hline
SphereFace     &  $\cos(m\theta_{y_i}) = \cos\theta_j$\\ \hline
CosFace        &  $\cos\theta_{y_i}-m = \cos\theta_j$ \\\hline
ArcFace        &  $\cos(\theta_{y_i}+m) = \cos\theta_j$\\ \hline
MV-Arc-Softmax & $\cos(\theta_{y_i}+m) = \cos\theta_j$ (easy)\\
              &  $\cos(\theta_{y_i}+m) = t\cos\theta_j+t-1 $ (hard)\\ \hline
\bf{CurricularFace (Ours)}           &  $\cos(\theta_{y_i}+m) = \cos\theta_j$ (easy)\\
              &  $\cos(\theta_{y_i}+m) = (t+\cos\theta_j)\cos\theta_j$(hard)\\\hline
\end{tabular}
}
\figvspace
\end{center}
\end{table}

\subsection{Discussions with SOTA Loss Functions}
\begin{figure}[t]
 \centering
 \includegraphics[trim={0 0 0 0mm},clip,width=\linewidth]{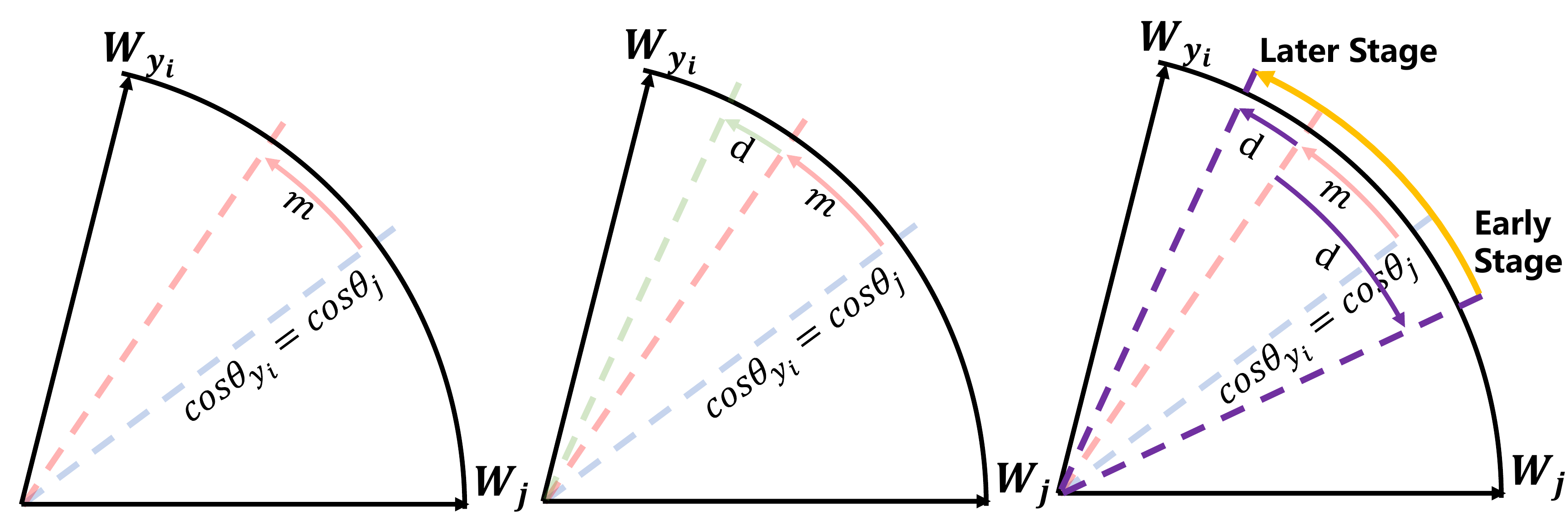}
 \caption{\small 
 \textbf{Blue line}, \textbf{red line}, \textbf{green line} and \textbf{purple line} denote the decision boundary of Softmax, ArcFace, MV-Arc-Softmax, and ours, respectively. $m$ denotes the angular margin added by ArcFace. $d$ denotes the additional margin of MV-Arc-Softmax and ours. In MV-Arc-Softmax, $d=(t-1)\cos\theta_j+t-1$. In ours, $d=(t+\cos\theta_j-1)\cos\theta_j$.
 }
 \label{fig:loss_comp}
 \figvspace
\end{figure}

\paragraph{Comparison with ArcFace and MV-Arc-Softmax.}
We first discuss the difference between our CurricularFace and the two competitors, ArcFace and MV-Arc-Softmax, from the perspective of the decision boundary in Tab.~\ref{tab:decision_boundary}.
ArcFace introduces a margin function $T(\cos\theta_{y_i})=\cos(\theta_{y_i}+m)$ from the perspective of the positive cosine similarity.
As shown in Fig.~\ref{fig:loss_comp}, its decision condition changes from $\cos\theta_{y_{i}} = \cos\theta_j$ (\textit{i.e.}, blue line) to $\cos(\theta_{y_i}+m) = \cos\theta_j$ (red line) for each sample.
MV-Arc-Softmax introduces additional margin from the perspective of negative cosine similarity for hard samples, and the decision boundary becomes $\cos(\theta_{y_i}+m)= t\cos\theta_j+t-1$ (green line).
Conversely, we adaptively adjust the weights of hard samples in different training stages.
The decision condition becomes $\cos(\theta_{y_i}+m)=(t+\cos\theta_j)\cos\theta_j$ (purple line).
During training, the decision boundary for hard samples changes from one purple line (\textbf{early stage}) to another (\textbf{later stage}), which emphasizes easy samples first and hard samples later.

\vspace{-3mm}
\paragraph{Comparison with Focal Loss.}
Focal loss 
is formulated as: $G(p(x))=\alpha(1-p(x_i))^\beta$,
where $\alpha$ and $\beta$ are modulating factors to be tuned manually.
The definition of hard samples in Focal loss is ambiguous, since it focuses on \textit{relatively} hard samples by reducing the weight of easier samples during entire training process.
In contrast, the definition of hard samples in our CurricularFace is more clear, \textit{i.e.}, mis-classified samples.
Meanwhile, the weights of hard samples are \textit{adaptively} determined in different training stages.


\section{Experiments}
\subsection{Implementation Details}
\paragraph{Datasets.}
We separately employ CASIA-WebFace~\cite{Yi2014learning} and refined MS$1$MV$2$~\cite{deng2018arcface} as our training data for fair comparisons with other methods. CASIA-WebFace contains about $0.5$M of $10$ individuals, and MS$1$MV$2$ contains about $5.8$M images of $85$K individuals.
We extensively test our method on several popular benchmarks, including LFW~\cite{LFWTech}, CFP-FP~\cite{sengupta2016frontal}, CPLFW~\cite{CPLFWTech}, AgeDB~\cite{moschoglou2017agedb}, CALFW~\cite{zheng2017crossage},  
IJB-B~\cite{whitelam2017iarpa}, IJB-C~\cite{maze2018iarpa}, and MegaFace~\cite{kemelmacher2016megaface}.

\paragraph{Training Setting.}
We follow~\cite{deng2018arcface} to crop the $112\times 112$ faces with five landmarks~\cite{zhang2016mtcnn,tai2019towards}.
For the embedding network, we adopt ResNet$50$ and ResNet$100$ as in~\cite{deng2018arcface}.
Our framework is implemented in Pytorch~\cite{paszke2017automatic}.
We train models on $4$ NVIDIA Tesla P$40$ GPU with batch size $512$. The models are trained with SGD algorithm, with momentum $0.9$ and weight decay $5e-4$. On CASIA-WebFace, the learning rate starts from $0.1$ and is divided by $10$ at $28$, $38$, $46$ epochs.
The training process is finished at $50$ epochs.
On MS$1$MV$2$, we divide the learning rate at $10$, $18$, $22$ epochs and finish at $24$ epochs. We follow the common setting as~\cite{deng2018arcface} to set scale $s=64$ and margin $m=0.5$ 
.

\begin{table}[t!]
\begin{center}
\small
\caption{\small \textbf{Verification performance (\%) of different values of $t$}.}
\label{tab:comp_different_values_of_t}
\begin{tabular}{l|cc}
\hline
Methods (\%)           & LFW  & CFP-FP  \\ \hline\hline
$t=0$               & $99.32$ &  $95.90$ \\
$t=0.3$             & $99.37$ &  $96.47$ \\
$t=0.7$             & $99.42$ &  $96.66$ \\
$t=1$               & $99.45$ &  $93.94$ \\\hline
Adaptive $t$        & $\bf99.47$ &  $\bf96.96$ \\\hline
\end{tabular}
\end{center}
\end{table}

\begin{table}[t!]
\begin{center}
\small
\caption{\small \textbf{Verification performance (\%) of different strategies for setting $t$}.}
\label{tab:comp_different_strategies_of_t}
\begin{tabular}{l|ccc}
\hline
Methods (\%)           & LFW  & CFP-FP  \\ \hline\hline
$\text{Mode}(\cos\theta_{y_i})$   & $99.42$ & $96.49$ \\
$\text{Mean}(p_{x_i})$               & $99.42$ &  $95.39$ \\
$\text{Mean}(\cos\theta_{y_i})$   & $\bf{99.47}$ &  $\bf{96.96}$\\\hline
\end{tabular}
\end{center}
\end{table}

\begin{figure}[t]
  \centering
  \includegraphics[trim={0 0 0 0mm},clip,width=0.9\linewidth]{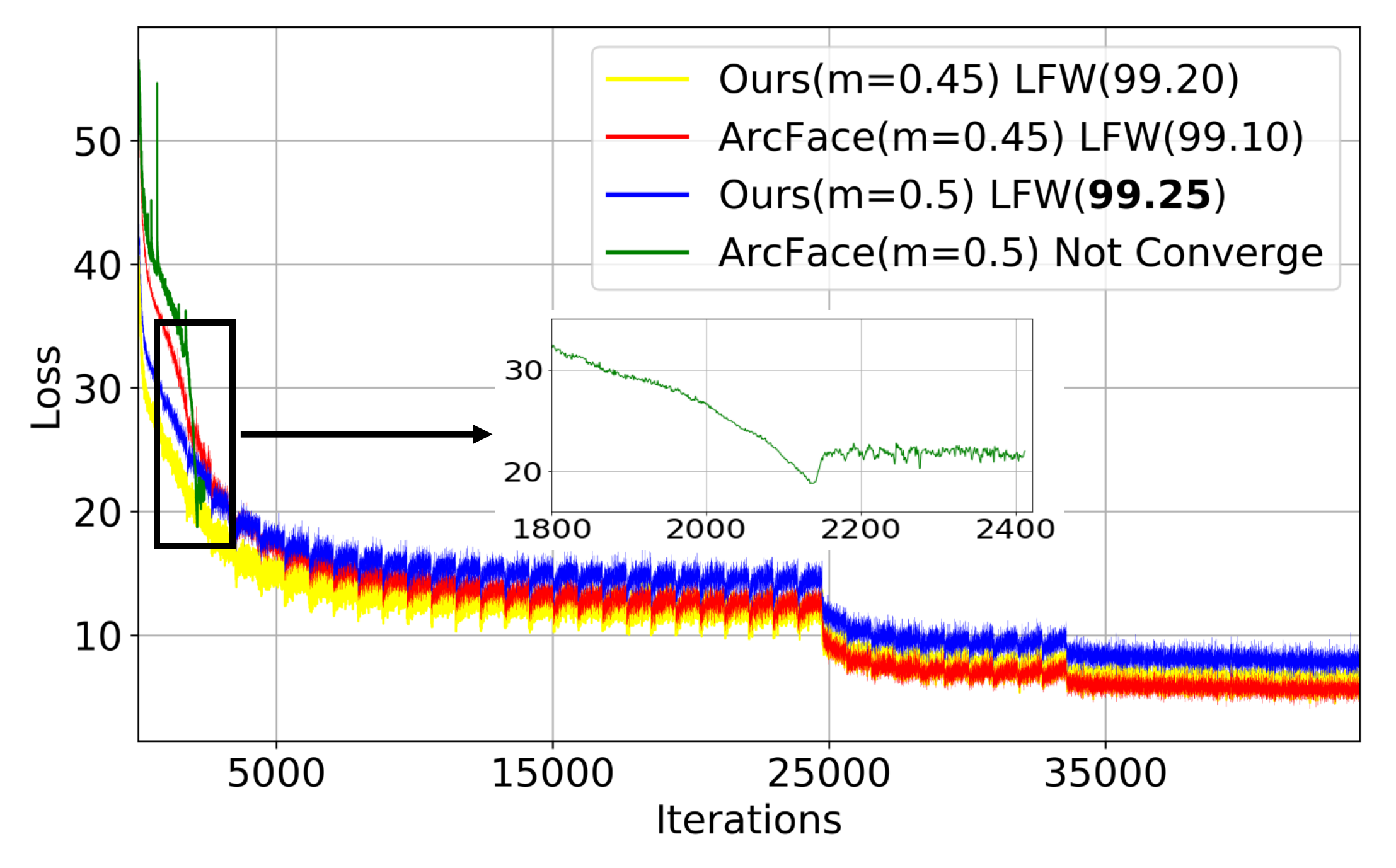}
  \caption{\small
  \textbf{Illustrations on loss curves} of our CurricularFace and ArcFace with the small backbone MobileFaceNet.
  }
  \label{fig:small_backbone}
\end{figure}

\subsection{Ablation study}
\paragraph{Effects on Fixed vs.~Adaptive Parameter $t$.}
We first investigate the effect of adaptive estimation of $t$.
We choose four fixed values between $0$ and $1$ for comparison.
Specifically, $0$ means the modulation coefficient $I(\cdot)$ of each hard sample's negative cosine similarity is always reduced based on its difficultness.
In contrast, $1$ means the hard samples are always emphasized.
$0.3$ and $0.7$ are between the two cases.
Tab.~\ref{tab:comp_different_values_of_t} shows that it is more effective to learn from easier samples first and hard samples later based on our adaptively estimated parameter $t$.

\paragraph{Effects on Different Statistics for Estimating $t$.}
We now investigate the effects of several other statistics, \textit{i.e.}, mode of positive cosine similarities in a mini-batch, or mean of the predicted ground truth probability for estimating $t$ in our loss.
As Tab.~\ref{tab:comp_different_strategies_of_t} shows: 
$1$) The mean of positive cosine similarities is better than mode.
$2$) The positive cosine similarity is more accurate than the predicted ground truth probability to indicate the training stages.

\vspace{-3mm}
\paragraph{Robustness on Training Convergence.}
As claimed in~\cite{li2019airface}, ArcFace exhibits the divergence issue when using small backbones like MobileFaceNet.
As a result, softmax loss must be incorporated for pre-training. 
To illustrate the robustness of our loss function on convergence issue with small backbones, we use the MobileFaceNet as the network architecture and train it on CASIA-WebFace.
As shown in Fig.~\ref{fig:small_backbone}, when the margin $m$ is set to $0.5$, the model trained with our loss achieves $99.25\%$ accuracy on LFW, while the model trained with ArcFace does not converge and the loss is \textbf{NAN} at about $2,400$-th step.
When the margin $m$ is set to $0.45$, both losses can converge, but our loss achieves better performance ($99.20\%$ vs.~$99.10\%$).
Comparing the yellow and red curves, since the losses of hard samples are reduced in early training stages, our loss converges much faster in the beginning, leading to lower loss than ArcFace. Later on, the value of our loss is slightly larger than ArcFace, because we emphasize the hard samples in later stages.
The results illustrate that learning from easy samples first and hard samples later is beneficial to model convergence.

\begin{table}[t!]
\begin{center}
\scriptsize
\caption{\small \textbf{Verification comparison with SOTA methods} on LFW, two pose benchmarks: CFP-FP and CPLFW, and two age benchmarks: AgeDB and CALFW. $*$ denotes our re-implemented results with the backbone ResNet100~\cite{deng2018arcface}.}
\label{tab:comp_pose}
\resizebox{1\columnwidth}{!}{
\begin{tabular}{l|ccccc}
\hline
Methods (\%) & LFW & CFP-FP & CPLFW & AgeDB & CALFW\\ \hline\hline
Center Loss (ECCV'$16$) & $98.75$ & $-$ & $77.48$  & $-$ & $85.48$ \\
SphereFace (CVPR'$17$) & $99.27$ & $-$ & $81.40$ & $-$ & $90.30$\\
DRGAN (CVPR'$17$) & $-$ & $93.41$ & $-$ & $-$ & $-$ \\
Peng~\textit{et al.} (ICCV'$17$) & $-$ & $93.76$ & $-$ & $-$ & $-$ \\
VGGFace2 (FG'$18$) & $99.43$ & $-$ & $84.00$ & $-$ & $90.57$\\
Dream (CVPR'$18$) & $-$ & $93.98$ & $-$ & $-$ & $-$ \\
Deng~\textit{et al.} (CVPR'$18$) & $99.60$ & $94.05$ & $-$ & $-$ & $-$ \\
ArcFace (CVPR'$19$) & ${99.77}$ & $98.27$ & $92.08$ & $98.15$ & $95.45$ \\
MV-Arc-Softmax (AAAI'$20$) & ${99.78}$ & - & - & $-$ & $-$ \\
MV-Arc-Softmax$*$  &  $\bf{99.80}$ & $98.28$ &$92.83$ & $97.95$ & $96.10$\\\hline
\textbf{CurricularFace (Ours)}  & $\bf{99.80}$ & $\bf{98.37}$ & $\bf{93.13}$  & $\bf{98.32}$ & $\bf{96.20}$ \\\hline
\end{tabular}
}
\end{center}
\end{table}

\begin{table}[t!]
\begin{center}
\scriptsize
\caption{\small \textbf{$\mathbf{1}$:$\mathbf{1}$ verification TAR (@FAR=$1$$e$$-4$)} on the IJB-B and IJB-C datasets. $*$ denotes our re-implemented results with the backbone ResNet100~\cite{deng2018arcface}.}
\label{tab:comp_ijb}
\begin{tabular}{l|cc}
\hline
Methods  (\%)          & IJB-B  & IJB-C  \\ \hline\hline
ResNet50+SENet50 (FG'$18$)    & $80.0$ &  $84.1$ \\
Multicolumn (BMVC'$18$)       & $83.1$ &  $86.2$ \\
DCN         (ECCV'$18$)       & $84.9$ &  $88.5$ \\
ArcFace-VGG2-R50 (CVPR'$19$)  & $89.8$ &  $92.1$ \\
ArcFace-MS1MV2-R100 (CVPR'$19$) & $94.2$ & $95.6$ \\
Adocos (CVPR'$19$)            &  $-$   &  $92.4$  \\
P2SGrad (CVPR'$19$)           &  $-$   &  $92.3$  \\
PFE (ICCV'$19$)               &  $-$   &  $93.3$  \\
MV-Arc-Softmax$*$ (AAAI'$20$) & $93.6$ &  $95.2$ \\\hline
Ours-MS1MV2-R100  & $\bf{94.8}$ & $\bf{96.1}$ \\\hline
\end{tabular}
\vspace{-3mm}
\end{center}
\end{table}

\subsection{Comparisons with SOTA Methods}
\paragraph{Results on LFW, CFP-FP, CPLFW, AgeDB and CALFW.}
Next, we train our CurricularFace on dataset MS$1$MV$2$ with ResNet$100$, and compare with the SOTA competitors on various benchmarks, including LFW for unconstrained face verification, CFP-FP and CPLFW for large pose variations, AgeDB and CALFW for age variations.
As reported in Tab.~\ref{tab:comp_pose}, our CurricularFace achieves comparable result (\textit{i.e.}, $99.80\%$) with the competitors on LFW where the performance is near saturated.
While for both CFP-FP and CPLFW, our method shows superiority over the baselines including general methods, \textit{e.g.},~\cite{wen2016discriminative},~\cite{cao2018vggface2}, and cross-pose methods, \textit{e.g.},~\cite{tran2017disentangled},~\cite{peng2017rec},~\cite{cao2018pose} and~\cite{deng2018uvgan}.
As a recent face recognition method, MV-Arc-Softmax achieves better performance than ArcFace, but still worse than Our CurricularFace.
Finally, for AgeDB and CALFW, 
as Tab.~\ref{tab:comp_pose} shows, our CurricularFace again achieves the best performance than all of the other SOTA methods.

\begin{figure}[t]
\centering
\subfigure[ROC for IJB-B]{
\begin{minipage}[t]{0.49\linewidth}
\centering
\label{fig:ijb_example1}
\includegraphics[width=1\linewidth]{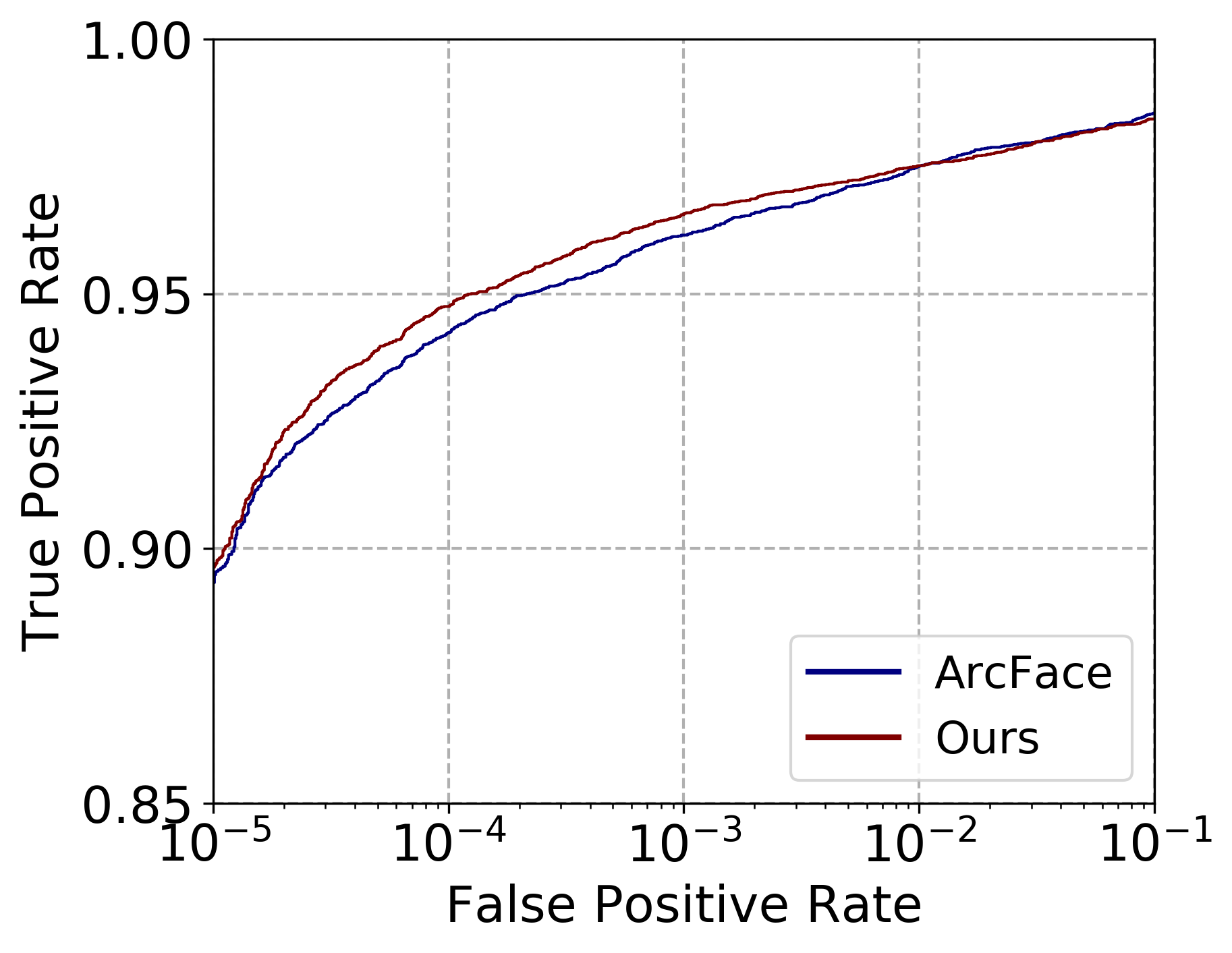}
\end{minipage}%
}%
\subfigure[ROC for IJB-C]{
\begin{minipage}[t]{0.49\linewidth}
\centering
\label{fig:ijb_example2}
\includegraphics[width=1\linewidth]{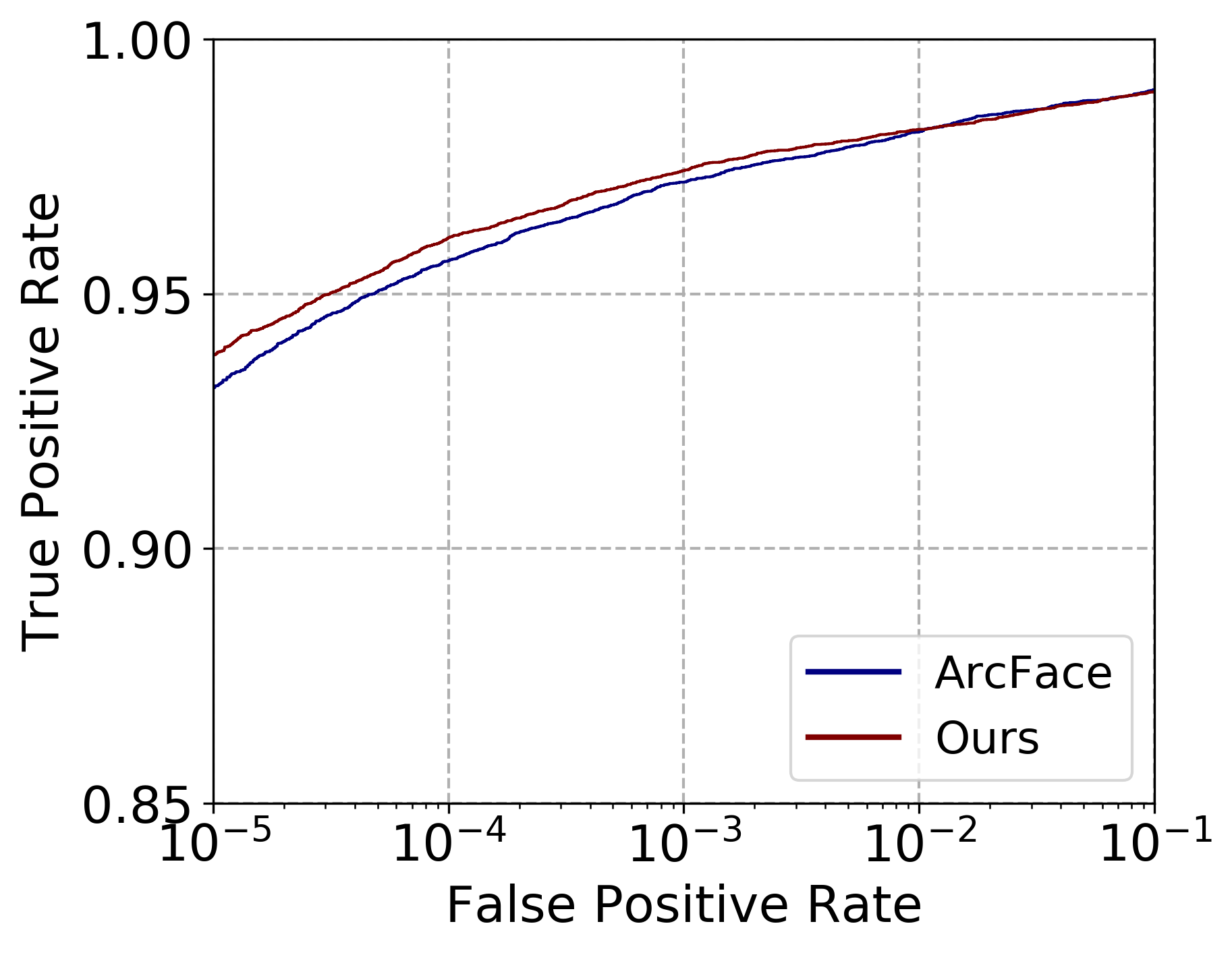}
\end{minipage}%
}%
\centering
\caption{\small \textbf{ROC of $\mathbf{1}$:$\mathbf{1}$ verification protocol} on IJB-B and IJB-C.
  }
 \label{fig:ijb_roc}
\end{figure}

\begin{figure*}[t]
  \centering
  \includegraphics[trim={0 0 0 0mm},clip,width=0.9\linewidth]{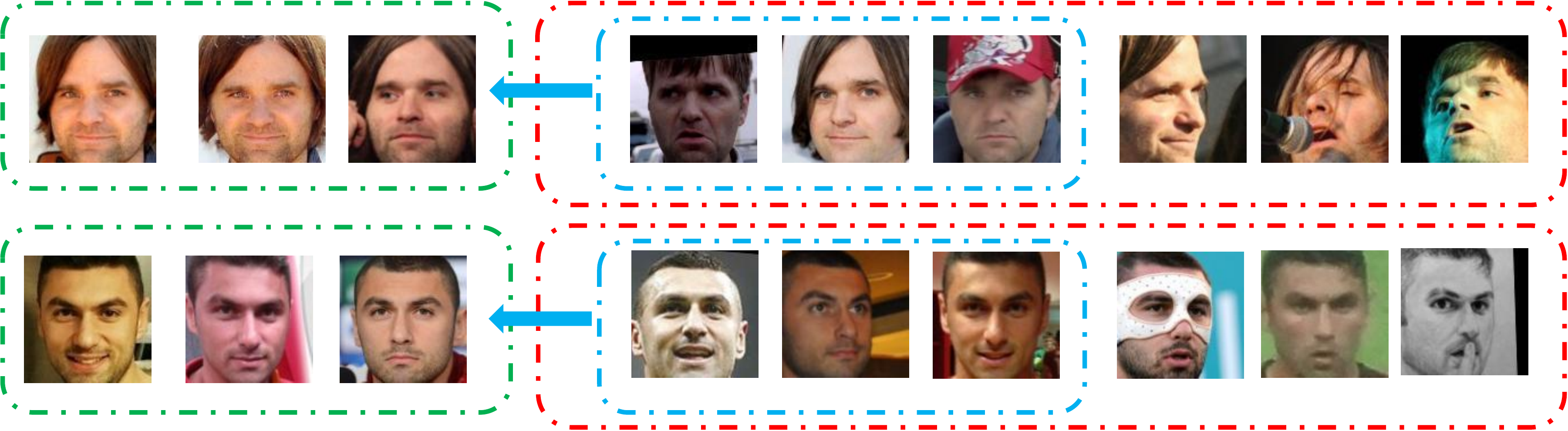}
  \caption{\small \textbf{Easy and hard examples from two subjects classified by our CurricularFace on early and later training stage, respectively.}
  Green box indicates easy samples.
  Red box indicates hard samples.
  Blue box means samples are classified as hard in early stage but re-labeled as easy in later stage, which indicates samples' transformation from hard to easy during the training procedure.
  }
  \label{fig:easy_hard_example}
  \figvspace
\end{figure*}

\vspace{-2mm}
\paragraph{Results on IJB-B and IJB-C.}
The IJB-B dataset contains $1,845$ subjects with $21.8$K still images and $55$K frames from $7,011$ videos. In the $1$:$1$ verification, there are $10,270$ positive matches and $8$M negative matches.
The IJB-C dataset is a further extension of IJB-B, which contains about $3,500$ identities with a total of $31,334$ images and $117,542$ unconstrained video frames.
In the $1$:$1$ verification, there are $19,557$ positive matches and $15,638,932$ negative matches.
On IJB-B and IJB-C datasets, we employ MS$1$MV$2$ and the ResNet$100$ for a fair comparison with recent methods. We follow the testing protocol in ArcFace and take the \textit{average of the image features} as the corresponding template representation without bells and whistles. Note that our method is not proposed for set-based face recognition task, and
DOES not adopt any specific strategies for set-based face recognition. The experiments on these two datasets are just to prove that our loss can obtain more discriminate features than the baselines like ArcFace, which are also generic methods for face recognition.
Tab.~\ref{tab:comp_ijb} exhibits the performance of different methods, \textit{e.g.}, Multicolumn~\cite{xie2018multicolumn}, DCN~\cite{xie2018comparator}, Adacos~\cite{zhang2019adacos}, P2SGrad~\cite{zhang2019p2sgrad}, PFE~\cite{shi2019probabilistic} and MV-Arc-Softmax~\cite{wang2018support} on IJB-B and IJB-C $1$:$1$ verification,
our method again achieves the best performance. Fig.~\ref{fig:ijb_roc} shows the ROC curves of CurricularFace and ArcFace on IJB-B/C with the backbone ResNet100, our method achieves better performance.

\begin{table}[t!]
\begin{center}
\scriptsize
\caption{\small \textbf{Verification comparison with SOTA methods} on MegaFace Challenge $1$ using FaceScrub as the probe set. “Id” refers to the rank-$1$ face identification accuracy with $1$M distractors, and “Ver” refers to the face verification TAR at $1e^{-6}$ FAR. The column “R” refers to data refinement on both probe set and $1$M distractors. $*$ denotes our re-implemented results with the backbone ResNet100~\cite{deng2018arcface}.}
\label{tab:comp_megaface}
\begin{tabular}{l|cccc}
\hline
Methods (\%) & Protocol & R & Id  & Ver  \\ \hline\hline
Triplet (CVPR'$15$)     & Small &            & $64.79$ & $78.32$ \\
Center Loss (ECCV'$16$) & Small &            & $65.49$ & $80.14$ \\
SphereFace (CVPR'$17$)  & Small &            & $72.73$ & $85.56$ \\
CosFace (CVRP'$18$)     & Small &            & $77.11$ & $89.88$ \\
AM-Softmax (SPL'$18$)   & Small &            & $72.47$ & $84.44$ \\
ArcFace-R50 (CVPR'$19$) & Small &            & $77.50$ & $92.34$ \\
ArcFace-R50             & Small & \checkmark & $91.75$ & $93.69$ \\\hline
Ours-R50                & Small &            & $\bf{77.65}$ & $\bf{92.91}$ \\
Ours-R50                & Small & \checkmark & $\bf{92.48}$ & $\bf{94.55}$ \\\hline\hline
CosFace-R100            & Large &            & $80.56$ & $96.56$ \\
CosFace-R100            & Large & \checkmark & $97.91$ & $97.91$ \\
ArcFace-R100            & Large &            & $81.03$ & $96.98$ \\
ArcFace-R100            & Large & \checkmark & $98.35$ & $98.48$ \\
PFE (ICCV'$19$)         & Large &            & $78.95$ & $92.51$  \\
Adacos (CVPR'$19$)       & Large & \checkmark & $97.41$ & $-$     \\
P2SGrad (CVPR'$19$)      & Large & \checkmark & $97.25$ & $-$     \\
MV-Arc-Softmax (AAAI'$20$) & Large & \checkmark & $97.14$ & $97.57$ \\
MV-Arc-Softmax*           & Large &            & $80.59$ & $96.22$ \\
MV-Arc-Softmax*           & Large & \checkmark & $97.76$ & $97.80$ \\\hline
Ours-R100                 & Large &            & $\bf{81.26}$ & $\bf{97.26}$ \\
Ours-R100                 & Large & \checkmark & $\bf{98.71}$ & $\bf{98.64}$ \\\hline\hline
AdaptiveFace-R50 (CVPR‘$19$)  & Large & \checkmark & $95.02$  & $95.61$ \\
Ours-R50                     & Large & \checkmark & $\bf{98.25}$  & $\bf{98.44}$ \\\hline

\end{tabular}
\end{center}
\end{table}

\begin{figure}[t!]
  \centering
  \includegraphics[trim={0 0 0 0mm},clip,width=0.8\linewidth]{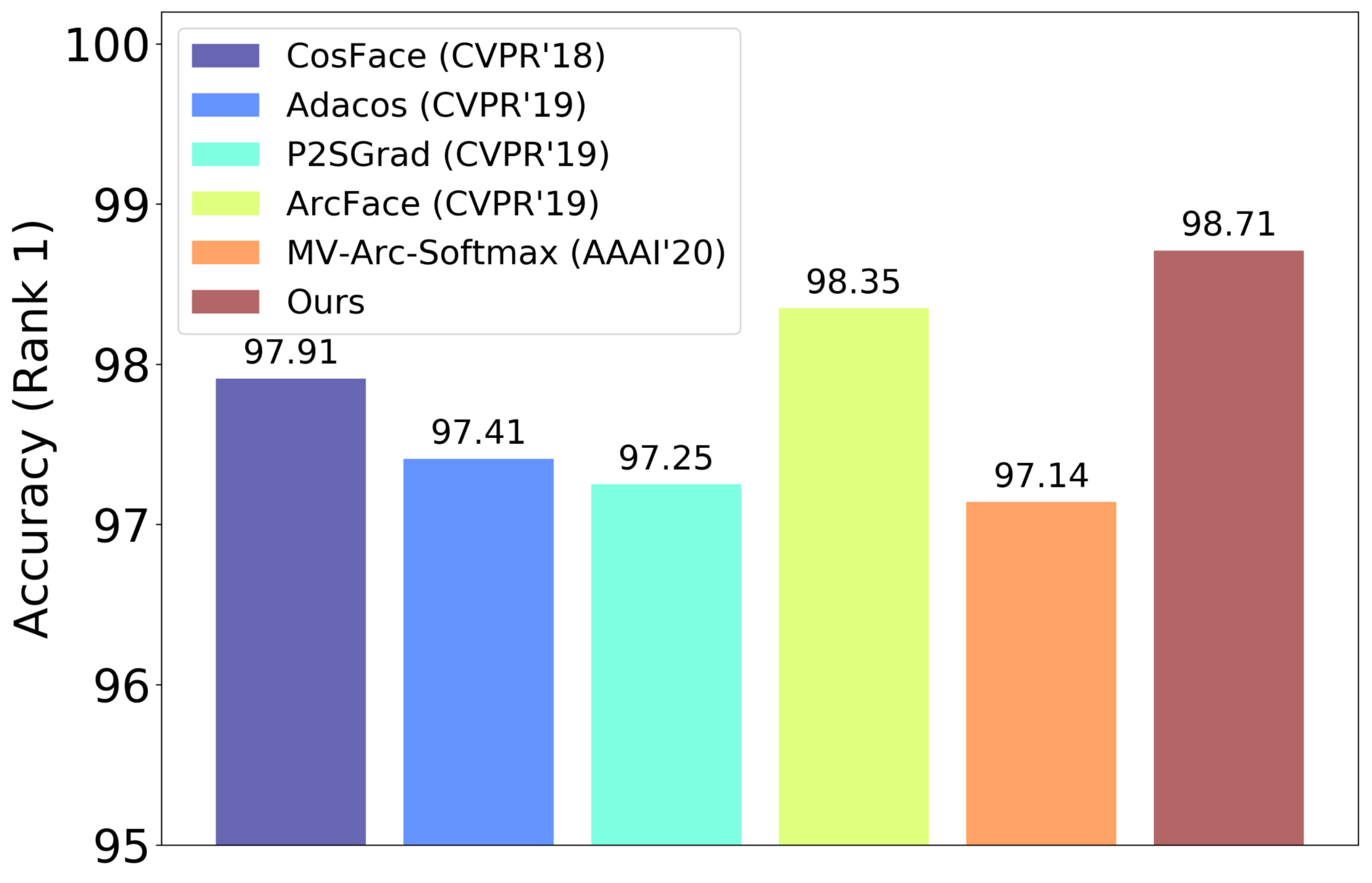}
  \caption{\small \textbf{The rank-$1$ face identification accuracy on MegaFace Challenge $1$} with both the $1$M distractors and the probe set refined by ArcFace.
  }
  \label{fig:megface_result}
  \figvspace
\end{figure}

\begin{figure}[t!]
\centering
\subfigure[TOP $1$]{
\begin{minipage}[t]{0.49\linewidth}
\centering
\label{fig:megaface_top1}
\includegraphics[width=1\linewidth]{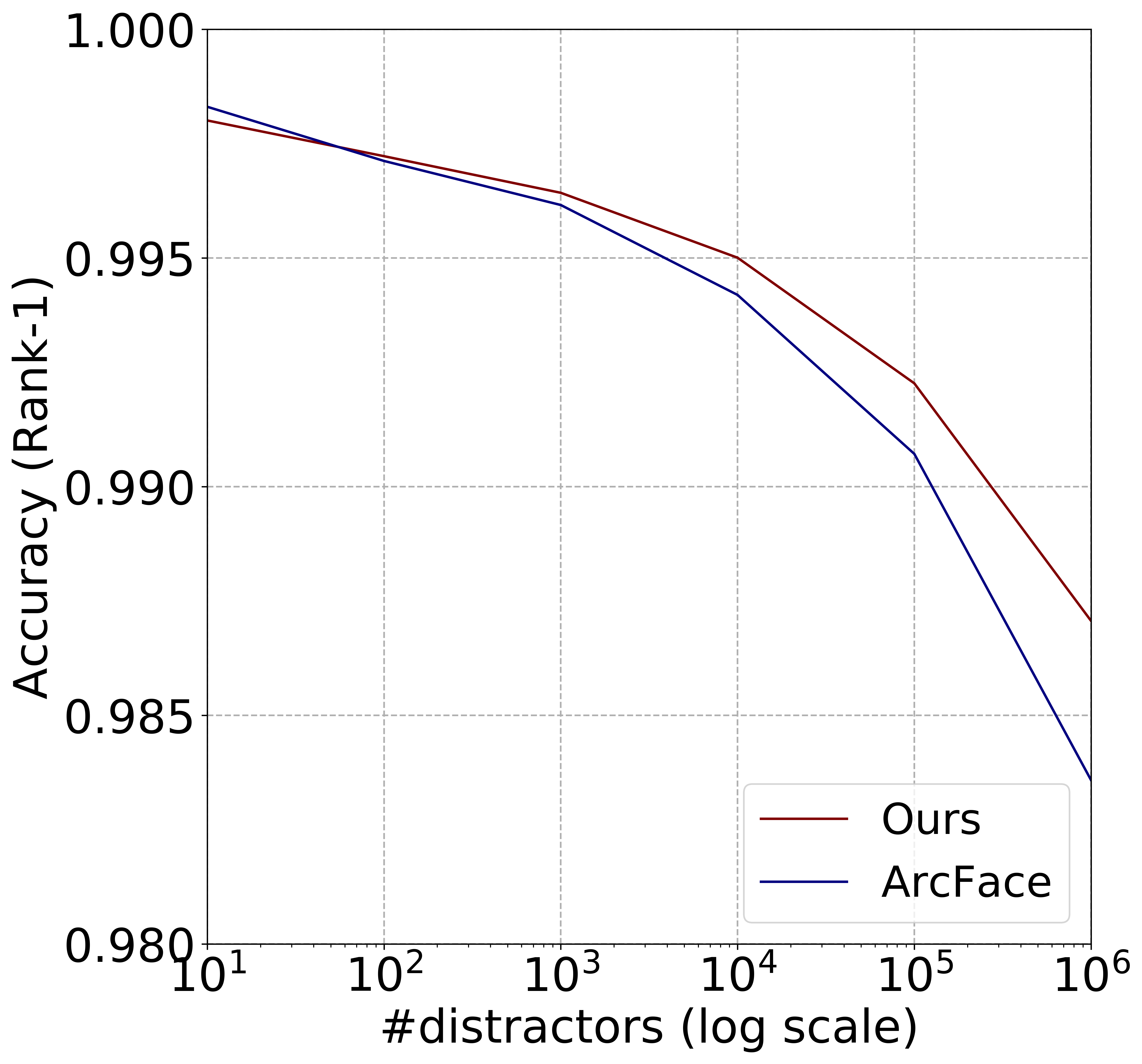}
\end{minipage}%
}%
\subfigure[ROC]{
\begin{minipage}[t]{0.49\linewidth}
\centering
\label{fig:roc_megaface}
\includegraphics[width=1\linewidth]{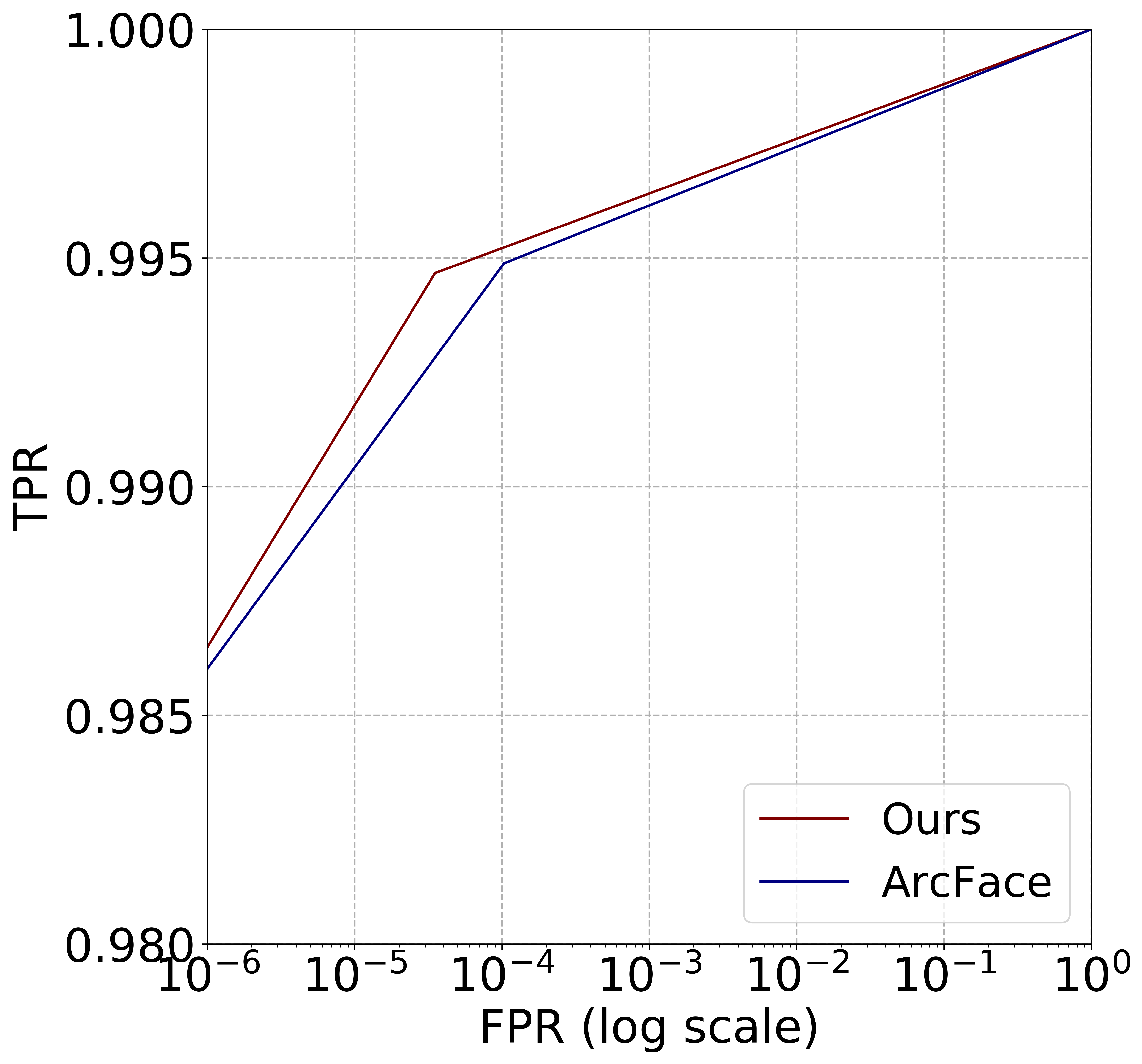}
\end{minipage}%
}%
\centering
\caption{\small \textbf{Illustrations on Top $1$ of different distractors and ROC on Megaface}. Results are evaluated on refined MegaFace dataset. The results of ArcFace are from the official ResNet$100$ pre-trained with MS$1$M.
  }
 \label{fig:megaface}\figvspace
\end{figure}

\paragraph{Results on MegaFace.}
Finally, we evaluate the performance on the MegaFace Challenge.
The gallery set of MegaFace includes $1$M images of $690$K subjects, and the probe set includes $100$K photos of $530$ unique individuals from FaceScrub.
We report the two testing results under two protocols (large or small training set).
Here, we use CASIA-WebFace and MS$1$MV$2$ under the small protocol and large protocol, respectively.
In Tab.~\ref{tab:comp_megaface}, our method achieves the best single-model identification and verification performance under both protocols, surpassing the recent strong competitors, \textit{e.g.}, CosFace, ArcFace, Adacos, P2SGrad and PFE.
We also report the results following the ArcFace testing protocol, which refines both the probe set and the gallery set.
As shown in Fig.~\ref{fig:megface_result}, our method still clearly outperforms the competitors and achieves the best performance on identification.
Compared with ArcFace, our loss shows better performance under both identification and verification scenarios as shown in Fig.~\ref{fig:megaface}. AdapitveFace~\cite{liu2019adaptiveface} is another recent margin-based loss function for face recognition. We train our model with the same training data MS1MV2 and the same backbone ResNet50~\cite{deng2018arcface} as AdaptiveFace for a fair comparison. The results in Tab.~\ref{tab:comp_megaface} demonstrate the superiority of our method.

\paragraph{Time Complexity.}
The proposed method only brings small burden on training complexity, but has the same cost as the backbone model during inference. Specifically, compared with the conventional margin-based loss functions, our loss only additionally adjusts the negative cosine similarity of hard samples. Under the same environment and batchsize, ArcFace~\cite{deng2018arcface} costs $0.370$s for each iteration on NVIDIA P40 GPUs, while ours costs $0.378$s.

\paragraph{Discussion on Easy and Hard Samples During Training.}
Finally, Fig.~\ref{fig:easy_hard_example} shows the easy and hard samples classified by our method in different training stages.
As we can see, the front and clear faces are usually considered as easy samples in early training stage, and our model mainly learns the identity information from these samples.
With the model continues training, slightly harder samples (\textit{i.e.}, Blue box) are gradually focused and corrected as the easy ones.

\section{Conclusions}
In this paper, we propose a novel Adaptive Curriculum Learning Loss that embeds the idea of adaptive curriculum learning into deep face recognition.
Our key idea is to address easy samples in the early training stage and hard ones in the later stage.
Our method is easy to implement and robust to converge.
Extensive experiments on popular facial benchmarks demonstrate the effectiveness of our method compared to the SOTA competitors.
Following the main idea of this work, future research can be expanded in various aspects, including designing a better function $N(\cdot)$ for negative cosine similarity that shares similar \textit{adaptive} characteristic during training, and investigating the effects of \textit{noise} samples that might be optimized as hard samples.

{\small
\bibliographystyle{ieee_fullname}
\bibliography{egbib}
}

\end{document}